\documentclass{article}

\usepackage{microtype}
\usepackage{graphicx}
\usepackage{subfigure}
\usepackage{booktabs} 

\usepackage{hyperref}

\usepackage[accepted]{synsml2023}

\usepackage{amsmath}
\usepackage{amssymb}
\usepackage{mathtools}
\usepackage{amsthm}

\usepackage[capitalize,noabbrev]{cleveref}

\theoremstyle{plain}

\theoremstyle{definition}

\theoremstyle{remark}

\usepackage[textsize=tiny]{todonotes}

\synsmltitlerunning{Combining Thermodynamics-based model of the centrifugal compressors and Active Machine Learning for Design Optimization}

\begin{document}

\twocolumn[
\synsmltitle{Combining Thermodynamics-based Model of the Centrifugal Compressors and Active Machine Learning for Enhanced Industrial Design Optimization}



\synsmlsetsymbol{equal}{*}

\begin{synsmlauthorlist}
\synsmlauthor{Shadi Ghiasi}{comp}
\synsmlauthor{Guido Pazzi}{comp2}
\synsmlauthor{ Concettina Del Grosso}{comp2}
\synsmlauthor{Giovanni De Magistris}{comp}
\synsmlauthor{Giacomo Veneri}{comp}

\end{synsmlauthorlist}

\synsmlaffiliation{comp}{Artificial Intelligence Team - Baker Hughes, Florence, Italy}

\synsmlaffiliation{comp2}{Software Engineering Team - Baker Hughes, Florence, Italy}

\synsmlcorrespondingauthor{Shadi Ghiasi}{shadi.ghiasi@bakerhughes.com}

\synsmlkeywords{Active Learning, Industrial Machine Learning, Surrogate Modeling, Design Optimization}

\vskip 0.3in
]



\printAffiliationsAndNotice{}  
\begin{abstract}

The design process of centrifugal compressors requires applying an optimization process which is computationally expensive due to complex analytical equations underlying the compressor’s dynamical equations.
Although the regression surrogate models could drastically reduce the computational cost of such a process, the major challenge is the scarcity of data for training the surrogate model. 
Aiming to strategically exploit the labeled samples, we propose the \textit{ActiveCompDesign} framework in which
we combine a thermodynamics-based compressor model (i.e., our internal software for compressor design) and Gaussian Process- based surrogate model within a deployable Active Learning (AL) setting.
We first conduct experiments in an offline setting and further, extend it to an online AL framework where a real-time interaction with the thermodynamics-based compressor's model allows the deployment in production.
\textit{ActiveCompDesign} shows a significant performance improvement in surrogate modeling by leveraging on uncertainty-based query function of samples within the AL framework with respect to the random selection of data points. Moreover, our framework in production has reduced the total computational time of compressor's design optimization to around 46\% faster than relying on the internal thermodynamics-based simulator, achieving the same performance. 

\end{abstract}

\section{Introduction and Related Works}
\label{intro}
Centrifugal compressors' design is an extensive computational process as it requires the optimization of numerous design variables which are the starting point of software simulations of complex dynamical equations \cite{ju2021aerodynamic}.  

While engineering-powered software simulations are reliable solution to the end user due to the established technology, they have more complex formulations due to higher inter-dependency of system variables \cite{garg2010time}. 

Surrogate modeling with Machine Learning (ML) models for computer simulations enables reducing the computational cost and time required for a design simulation while maintaining a desired performance in industrial applications \cite{bicchi2022ai,owoyele2022application,kim2010surrogate}.
However, generating sufficient data points for training ML models is a daunting task since it requires running extensive software simulations. Therefore, without any strategic sampling, the possibility to explore a larger design space is limited.




Under such circumstances, the Active Learning (AL) strategy is a powerful framework to alleviate the problem of high quality annotation scarcity \cite{settles2012active}.
AL is a ML technique that allows the model to interact with an oracle by queering the most important data for learning \cite{monarch2021human}. In industrial applications, AL can make the most of resources by significantly reducing the amount of labeled data for training ML models \cite{brevault2022active}.

Utilizing ML surrogate models in the industrial simulation design setting has been explored by previous research. Kim et al. implement surrogate modeling for optimization of a centrifugal compressor impeller \cite{kim2010surrogate}. However, without any strategic sampling, the research is done for a limited design space. \textit{AutoML-GA} \cite{owoyele2022application} is an application of an automated machine learning-genetic algorithm coupled with computational fluid dynamics simulations for rapid engine design optimization. Chabanet et al. \cite{chabanet2021coupling} apply AL in Industry 4.0 context. Moreover, Murugesan et al. \cite{murugesan2022deep} propose an AL framework for estimating the operating point of a Modular Multi Pump used in energy field. Wang et al. \cite{wang2022active} apply AL for multilingual finger spelling corpora. Finally, see also \cite{reker2019practical} for some practical considerations for active ML in drug discovery. However, an AL based framework has not been explored for design optimization of centrifugal compressors. Moreover, most research have focused on offline evaluation of AL strategies, while, in industrial settings the data is acquired in real-time \cite{cacciarelli2022online} and a deployable streaming based AL framework is needed. 



In this study, we present the \textit{ActiveCompDesign} framework for deployable AL based design optimization of centrifugal compressors. 
We leverage on Gaussian Processes (GPs) as deep surrogates of centrifugal compressor dynamics coupled with the AL strategy with a design goal to reach the optimal power absorbed by the machine. We perform extensive computer simulations using our internal thermodynamics-based model integrated with an optimization algorithm to generate sufficient data samples for surrogate model training. We then use this data to perform an offline AL algorithm with GP surrogates as a proof of concept. We further deploy our framework through an online AL simulation environment in which the thermodynamics-based model and the ML-based model interact in real-time using a stream-based AL strategy. Our framework is currently in production. To the best of our knowledge, no other study have combined compressor's thermodynamics-based models and ML to propose a production-ready AI enhanced design optimization framework for design optimization of centrifugal compressors.



\section{\textit{ActiveCompDesign} framework}

\subsection{Problem definition}

We  consider the optimization of the boil off gas process of centrifugal compressors where the aim is to minimize the absorbed power by the machine. This is currently done based on a design optimization process on an internal thermodynamics-based simulator. Since the analytical equations underlying this physical process is complex and computationally expensive, we aim to strategically run the thermodynamics-based simulator during the optimization process with the minimum number of queries to reach the desired power.

We do this within the \textit{ActiveCompDesign} framework by integrating a regression surrogate model of the physical process throughout the optimization process to benefit from a faster calculation of the system's response. However, due to the uncertainties produced by the surrogate model, we want to rely on the actual physical process equations to obtain reliable outputs when the ML surrogate model produces high uncertainties in prediction.
In the following subsections we provide formulation for surrogate modeling of the design optimization process and the implemented offline and online AL framework. 

\subsection{Surrogate modeling and active learning}
Let $\mathbf{x} \subset \mathbb{R}^{d_{\text {in }}} $  be the input set where $d$ is the number of design parameters chosen by the expert engineer and $\mathbf{y} \subset \mathbb{R}^{d_{\text {out }}} $ is the desired output. 
A collection of $N$ number of runs of the thermodynamics-based simulator, will result in a finite number of pairs  $\mathcal {D}_{train}= \left(\mathbf{x}_n,\mathbf{y}_n \right) _{n=1}^N$ which is considered as the training data. These pairs represent sampled inputs and outputs of a complex analytical function $\mathbf{y}=f (\mathbf{x})$ which is encoded in the simulator. The aim of surrogate modeling is to estimate a function $\widehat{f}: \mathbb{R}^{d_{\text {in }}} \rightarrow \mathbb{R}^{d_{\text {out }}}$ which should be as close as possible to the true function $f$. We aim to condition the parameters of $\widehat{f} (\theta)$ through $\widehat{f}(x,\theta | \mathcal {D}_{train}) =\widehat{y}, \forall x \in \mathcal{B} : \widehat{y} \sim {y}$
where $\mathcal{B}$ is the bounded subspace in $\mathbb {R}$.
In design optimization we look for those values of $x$ leading to the minimum $y$. Therefore, by integrating the surrogate model into the optimization algorithm we aim to: 
\begin{equation}
 min|\sim {y}  \
    \text{subject to} \widehat{f}(x,\theta | \mathcal {D}_{train}) =\widehat{y}, \forall x \in \mathcal{B} : \widehat{y} \sim {y}
\end{equation}

In the \textit{ActiveCompDesign} framework, our design parameters ($x$) are a set of 12 flow coefficient rates corresponding to different compression stages. Each coefficient is bounded between a minimum and maximum value designed by the expert engineer. The output is the total absorbed power ($\mathcal{P}$) calculated by the thermodynamics-based simulator. Therefore, a single objective optimization algorithm is performed to reach a desired $\mathcal{P}$.

\subsection{Offline {\textit{ActiveCompDesign} framework}}

We gather data by running N number of runs of the thermodynamics-based model during an optimization process selected by the expert engineers to obtain a baseline minimum power. 
Given this collected dataset ($\mathcal {D}^{Start}_{t}$) comprising of a set of flow coefficents as $x$ and power as $\mathcal{P}$, we train a regressor $\widehat{f}^{Start}$ on a small pool of labeled data ($\mathcal{D}^{Pool}_{t}$). Throughout the AL process the regressor and the original dataset get updated resulting in $\widehat{f}_{Up}$ and $\mathcal{D}^{Up}_{t}$. 
The selection of the new observations to be labeled by the simulator to be considered for the new training dataset is obtained using query strategies formulated based on the regressor. 

We particularly choose GPs \cite{williams2006gaussian} as the surrogate regression model thanks to their high performance in mapping input-output relationship in our study and their Bayesian structure (A list of regression models tested in our data set and a comparison of their performance is in the Appendix).
A GP is defined by its mean function $\mu(x)$ and a Kernel function computing the covariance function between datapoints $K(x_{i},x_{j})$, therefore $\widehat{f} \sim \mathcal{GP} (\mu(x_{i}),K(x_{i},x_{j}))$.

Within the GP regression framework we are able to compute the posterior mean and posterior variance for the prediction of each sample. To perform AL with GP we select the next training sample based on the maximum variance  \cite{kapoor2007active,zhao2021efficient,yue2020active,zimmer2018safe}. 
For implementation of offline \textit{ActiveCompDesign} we rely on the \textit{ModAL} library \cite{danka2018modal}. 

\subsection{Online {\textit{ActiveCompDesign} framework}}
\begin{algorithm}[tb]
	\caption{Online {\textit{ActiveCompDesign} framework}}
	\label{alg:example}
	\begin{algorithmic}[1]
		\STATE {\bfseries Input:} Flow coefficients ($X$), Bounded space of flow coefficients ($\mathcal{B}$), Number of iterations for pre-training ($N_{PT}$), Number of total iterations ($N_{tot}$)
		\STATE {\bfseries Output:} Power absorbed by the machine ($\mathcal{P}$)
		\STATE Initialize $X$= arbitrary initial flow coefficient values. 
		\FOR{$i=1$ {\bfseries to} $N_{PT}$}
		\STATE Get the input ($X$) from the optimizer within  ($\mathcal{B}$). 
		\STATE Compute the power $\mathcal{P}$ based on thermodynamics-based model ($\mathcal{P}_{Phys}$).
		
		\STATE Create $\mathcal{D}^{Pretrain}$
		\STATE Train the pre-trained $\mathcal{GP}$ model ($\mathcal{GP}_{pt}$) on this training set.
		\STATE Compute the uncertainty threshold $Ub$ based on  $\mathcal{GP}$ posterior variances of previous samples.
	
		\ENDFOR
		\FOR {$i=N_{PT}+1$ {\bfseries to} $N_{tot}$}
		\STATE Get the input ($X$) from the optimizer within  ($\mathcal{B}$). 
		\STATE Compute the power calculated from GP ($\mathcal{P}_{GP}$), uncertainty $(\mathcal{P}_{GP})$ from $\mathcal{GP}_{pt}$.
		\IF{ Uncertainty $(\mathcal{P}_{GP}) < Ub$}
		\STATE Return $\mathcal{P}$=$\mathcal{P}_{GP}$
		\STATE Update $Ub$.
		\ELSE 
		\STATE{Return $\mathcal{P}$=$\mathcal{P}_{Phys}$.}
		\STATE{Update $\mathcal{GP}_{pt}$ with the new data point.}
     	\STATE Update $Ub$.
		\ENDIF
		\ENDFOR

	\end{algorithmic}
\end{algorithm}
With this approach we design a simulation environment where the thermodynamics-based simulator, the optimizer and the surrogate model can interact in real time after each data streaming.

The primary difference between the online and offline framework is that the labels of new observations can non longer be queried in a pool based way. With real time streaming of data points an instant decision should be made to whether label the data points or to discard the sample for learning. 

Since our optimization goal is to reach minimum power ($\mathcal{P}$) with less expensive computational effort but keeping the reliability of the output, we rely on the GP surrogate model only as an alternative model in case the uncertainty of the prediction is high. 
To this end, we consider a maximum number of iterations ($N_{PT}$) for the optimizer and the thermodynamics-based simulator to interact to generate an initial labeled dataset $\mathcal{D}^{Pretrain}=\left(\mathbf{x}_n,\mathbf{y}_n \right) _{n=1}^{N_{tot}}$. We then train a surrogate model based on GPs and consider the mean of the posterior variances of the predictions as the uncertainty threshold ($Ub$) for the next iteration.
We examine if this replicated simulation is able to achieve comparable performance in suggesting optimal design parameters for obtaining the minimum power. 
Our thermodynamics-based simulator and the code for the framework in this paper are proprietary. More details of the online framework is reported in algorithm \ref{alg:example}. 

\begin{figure*}[ht]
	\vskip 0.2in
	
	\begin{center}
		\centerline{\includegraphics[width=0.8\linewidth]{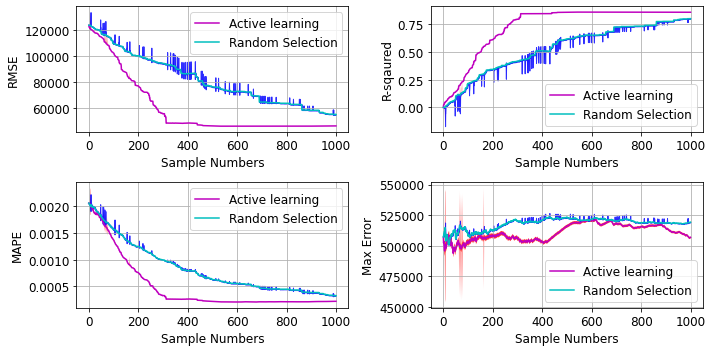}}
		\caption{Comparison of performance metrics of surrogate modeling with Gaussian Process regression model by selecting the samples with random acquisition with respect to active learning framework.  The vertical bars for each sample show the standard deviation of the performance metric as a result of uncertainty for     prediction.}
    \label{metrics}
	\end{center}
	\vskip -0.2in

\end{figure*}

\begin{table}[]
\caption{Computational time of the proposed framework compared to the baseline thermodynamics-based-simulator. N shows the number of iterations for pre-training and total number of runs are the number of iterations for \textit{ActiveCompDesign} to achieve the minimum baseline power.}
\label{online_res}
\begin{tabular}{llll}
	\toprule
   & \begin{tabular}[c]{@{}l@{}}Number \\ of runs\end{tabular} & \begin{tabular}[c]{@{}l@{}}Pre-training \\ time (s)\end{tabular} & \begin{tabular}[c]{@{}l@{}}Total \\ time (s)\end{tabular} \\ \hline
\begin{tabular}[c]{@{}l@{}} Thermodynamics  \\ model (baseline)\end{tabular}                                                           & 4000                                                      &      -                                                            & 109080                                                    \\ \hline
\begin{tabular}[c]{@{}l@{}}\textit{ActiveCompDesign} \\ ($N_{PT}$=50)\end{tabular}  & 200                                                       & 2340                                                             & 7180                                                      \\ \hline
\begin{tabular}[c]{@{}l@{}}\textit{ActiveCompDesign} \\ ($N_{PT}$=100)\end{tabular} & 160                                                       & 4680                                                             & 5040                                                      \\ \hline
\begin{tabular}[c]{@{}l@{}}\textit{ActiveCompDesign}\\  ($N_{PT}$=150)\end{tabular} & 120                                                       & 7020                                                             & 6280                       \\ \hline                               
\end{tabular}
\end{table}

\section{Results and Discussion}
\subsection{Offline {\textit{ActiveCompDesign} framework}}

To generate the dataset for training and evaluating the GP surrogate model, we perform a Bayesian optimization using GPs (from the \textit{Scikit-optimize library \cite{louppe2017bayesian}}) on top of our thermodynamics-based simulator of compressor design. 
This optimization process has led to 4000 runs in 12 dimensional input parameter space which constructs our labeled dataset. 

We select the surrogate GPs's hyper-parameters by performing a grid search hyper-parameter selection based on cross validation performance \cite{bergstra2011algorithms}. The best performing model is achieved with the \textit{Matern} kernel with length scale of 0.75 and $\nu$ of 0.5 (refer to \cite{williams2006gaussian} for formulation of the \textit{Matern} kernel). 

We perform model training with AL and compare the results with uniformly random acquisition of the samples as the baseline. Figure \ref{metrics} reports the regression performance metrics (i.e., Root Mean Squared Error (\textit{RMSE}), \textit{R-sqaured}, Mean Absolute Percentage Error (\textit{MAPE}) and \textit{Max error} as more samples are queried for training based on random selection and AL. We observe that the uncertainty-based query strategy leads to a much greater decrease of \textit{RMSE} and \textit{MAPE} metrics with less samples compared to the random sampling. This has lead to a clear improvement of error achieving almost the full-dataset performance by relying on only around 30\% of labeled data points for training. Moreover, the model's goodness of fit quantified by \textit{R-squared} has a faster increase using the AL strategy compared to the random selection.

As through GPs, we are able to obtain the standard deviation of prediction ($\sigma$), we also consider the regression performance for $y$+$\sigma$ and $y$-$\sigma$ where $y$ is the predicted value for each sample. As the plots in Figure \ref{metrics} show, there are high variations in regression performances with random selection, while with AL, this variation exists only for few initial samples and it disappears as the samples strategically grow in number.

\subsection{Online {\textit{ActiveCompDesign} framework}}
We evaluate the results of the deployed streaming-based AL framework based on the total number of runs required for achieving the baseline performance and the computational cost associated with it. In this regard, we consider different simulations where in each simulation $N_{PT}$ number of initial iterations is needed to pre-train the surrogate model. We compare these results in Table \ref{online_res}. The baseline model is the thermodynamics-based simulator where it achieved a desired minimum power through 4000 iterations lasting for around 30 hours. 
These results show a trade-off between the number of runs to interact with thermodynamics-based model and the total number of iterations for the whole framework to achieve the desired $\mathcal{P}$.
For the \textit{ActiveCompDesign} simulators, the highest computational cost is the pre-training cost where an interaction with oracle is needed. The simulator with 100 iterations for pre-training with a total number of 160 runs have taken the least total time of simulation with an improvement of around 46\% decrease in total computational time compared to the thermodynamics-based model. All our experiments have been performed using \textit{Nvidia- DGX1} with \textit{8x Tesla P100} GPUs and 20-core dual \textit{Intel} CPUs. 

\section{Conclusion}


The benefits of combining active ML methods with physical models underlying compressor's dynamics are large for design optimization applications including faster computations and more accurate design solutions. 
However, the trade-off between performance and computational power has to be carefully evaluated for the specific design application. Moreover (see also \cite{pardakhti2021practical}), the adoption of AL methods poses significant challenges in many practical applications, such as lack of data, discontinuous space of exploration and measurement error. 

The results obtained from offline and online \textit{ActiveCompDesign} show that integrating ML in compressor's simulators is viable for production ready application on the energy sector. Indeed, our framework is currently working in a production environment. For future research we will expand this framework for wider design optimization applications and improve the monitoring of the deployed model. 

\section{Impact Statement}
The inclusion of ML models into an internal thermodynamics-based model in our study offers a significant positive impact on various aspects of design optimization process in the turbo-machinery industry, including performance improvement, code optimization and enhancement of user experience. However, it is important to address the negative impacts such as privacy concerns and the potential inaccuracies in model prediction generated by dataset distribution shift and other factors.





\bibliography{calc_main.bib}
\bibliographystyle{synsml2023}

\newpage
\appendix
\onecolumn
\section{Appendix}

\subsection{General overview of the \textit{ActiveCompDesign} framework}

\begin{figure*}[ht]
	\vskip 0.2in
	\begin{center}
		\centerline{\includegraphics[width=15cm]{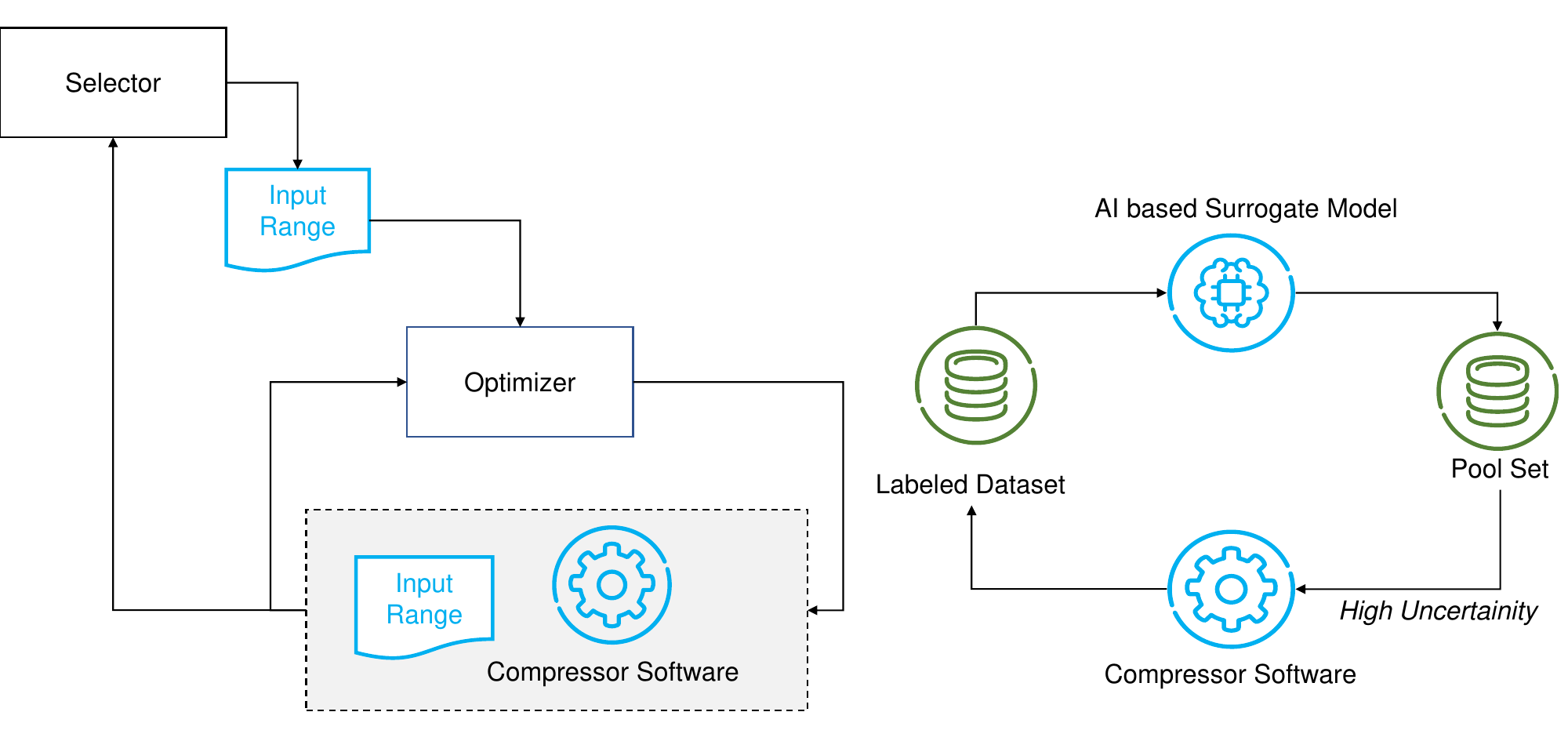}}
		\caption{General pipeline of \textit{ActiveCompDesign} framework.  }
		\label{calc_arch}
	\end{center}
	\vskip -0.2in
\end{figure*}

\subsection{Surrogate modeling selection}
In order to obtain an AI based surrogate model of the compressor's designer software, we rely on state-of-the-art supervised regression models to map the input-output relationship. Moreover, we select those regression models where we are able to obtain a quantification of the uncertainty of the prediction. 
Here a comparison of the performance of all tested models is listed \ref{surr_perf}. For random forests, gradient boosting and extra tree regressors we consider 100 numbers of estimators for training. For GPs we use a Matern kernel with parameters described in the text. We observe that GPs have achieved the best performance among all the models.

\begin{table*}[h]
\caption{Model development performance.}
\label{surr_perf}
	\vskip 0.15in

\begin{center}
\begin{small}
\begin{sc}

\begin{tabular}{lllllll}

\toprule
Model & \textit{RMSE} & \textit{Max Error} & \textit{R Squared} & \textit{MAPE}  \\
		
	\midrule
		
Random Forest    & 18937.12 & 112618.01  & 0.84 & 0.0003 \\ 
\hline
Gaussian Process & 4412.71  & 53286.78   & 0.99 & 0.00002 \\ 
\hline
Gradient Boosting  & 16919.41 & 50417.80   & 0.87 & 0.0002\\
\hline
Extra Tree Regressor   & 18413.44  & 106056.33  & 0.85 & 0.0003 \\ 
\hline
 
\end{tabular}
\end{sc}
\end{small}
\end{center}
\vskip -0.1in
\end{table*}


\end{document}